\documentclass[11pt]{article}
\usepackage[cp850]{inputenc}
\usepackage{amsmath, amsfonts}
\usepackage{graphicx}
\usepackage {xcolor}
\usepackage{hyperref}
\usepackage{cleveref}
\usepackage{cancel}

\usepackage{mathtools}

\newcommand{\verteq}{\rotatebox{90}{$\,=$}}
\newcommand{\equalto}[2]%
{\underset{\scriptstyle\overset{\mkern4mu\verteq}{#2}}{#1}}

\newcommand{\rmd}{\rm d}

\begin{document}
$\ $

\vspace{3cm}

\centerline{\Large{\bf ESGI 158 Report}}
\centerline{\Large{\bf Safe trajectory of a piece moved by a robot}}

\begin{center}
\today
\end{center} 
\vspace{0.5cm}

\noindent Ernest Benedito, ernest.benedito@upc.edu, \\
Oliver Bond, Bond@maths.ox.ac.uk,\\
Thomas Babb, Babb@maths.ox.ac.uk, \\
Juan R.~Pacha, juan.ramon.pacha@upc.edu,\\
Sandeep	Kumar, skumar@bcamath.org, \\
Joan Sol\`{a}-Morales, jc.sola-morales@upc.edu.

\vspace{0.5cm}

\section{Introduction}
\label{sec:Intro}
The company F.EE (\url{www.fee.de/en.html}) is an international company supplying automation technologies for industry, mostly in the automotive sector. The company is interested in determining the equations of the trajectories and the orientations of pieces that are moved by the action of a robot with seven degrees of freedom, of which six correspond to rotations of the arms and one to longitudinal transfers. In many cases, the piece is a curved metal sheet, and in the others, a three-dimensional object.

At the initial moment, the piece is at a certain point/location, orientation, and rest position, then the robot grabs and leaves it to another given point/location, orientation, and rest position corresponding to the other time instant. During its movement, the piece can suffer reversible and irreversible deformations caused by both mass forces and surface forces. The mass forces are due to translation and rotations (depending on accelerations and angular velocity of the body) while surface forces are due to air drag (depending on the velocity of the body) that act on it.
The company wants that the resulting motion does not result in irreversible deformations on the piece and that the time between the initial and final moments does not exceed a certain threshold. The actual movement of a piece is a consequence of the movements of the robot. This in turn depends on the design of the trajectory to be followed by the piece, as well as on the robot's ability to faithfully follow the desired trajectory. Any solution to the problem must consider not only the conditions under which the desired trajectory cannot cause irreversible deformation but also, the conditions under which the robot can follow the trajectory. The former depends on the desired velocity, acceleration, angular velocity of the piece, and the latter depends on the so-called {\sl jerk}, or, the first-derivative of the accelerations.

Certainly, the problem is complex and is rarely addressed in the scientific literature. There is abundant literature on the trajectory design of a piece moved by a robot with varying degrees of freedom \cite{whitney1972mathematics,braibant_geradin_1987} but the deformations caused by mass forces or friction are generally not taken into account.

This is certainly a matter of practical interest to the manufacturing industry and that it should be for the scientific community as well. 
\begin{figure}
\centering
\includegraphics[width=7cm]{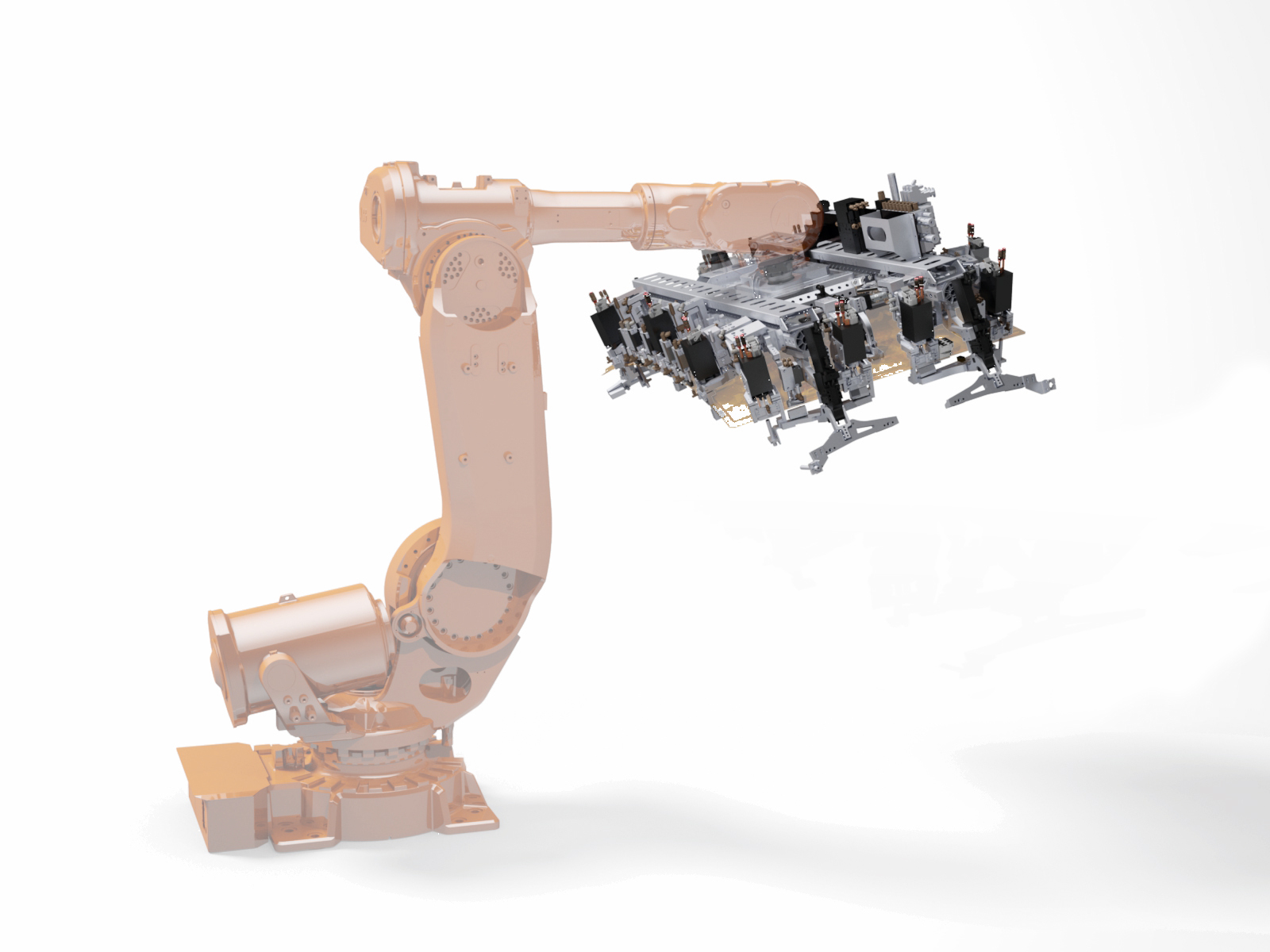}
\includegraphics[width=7cm]{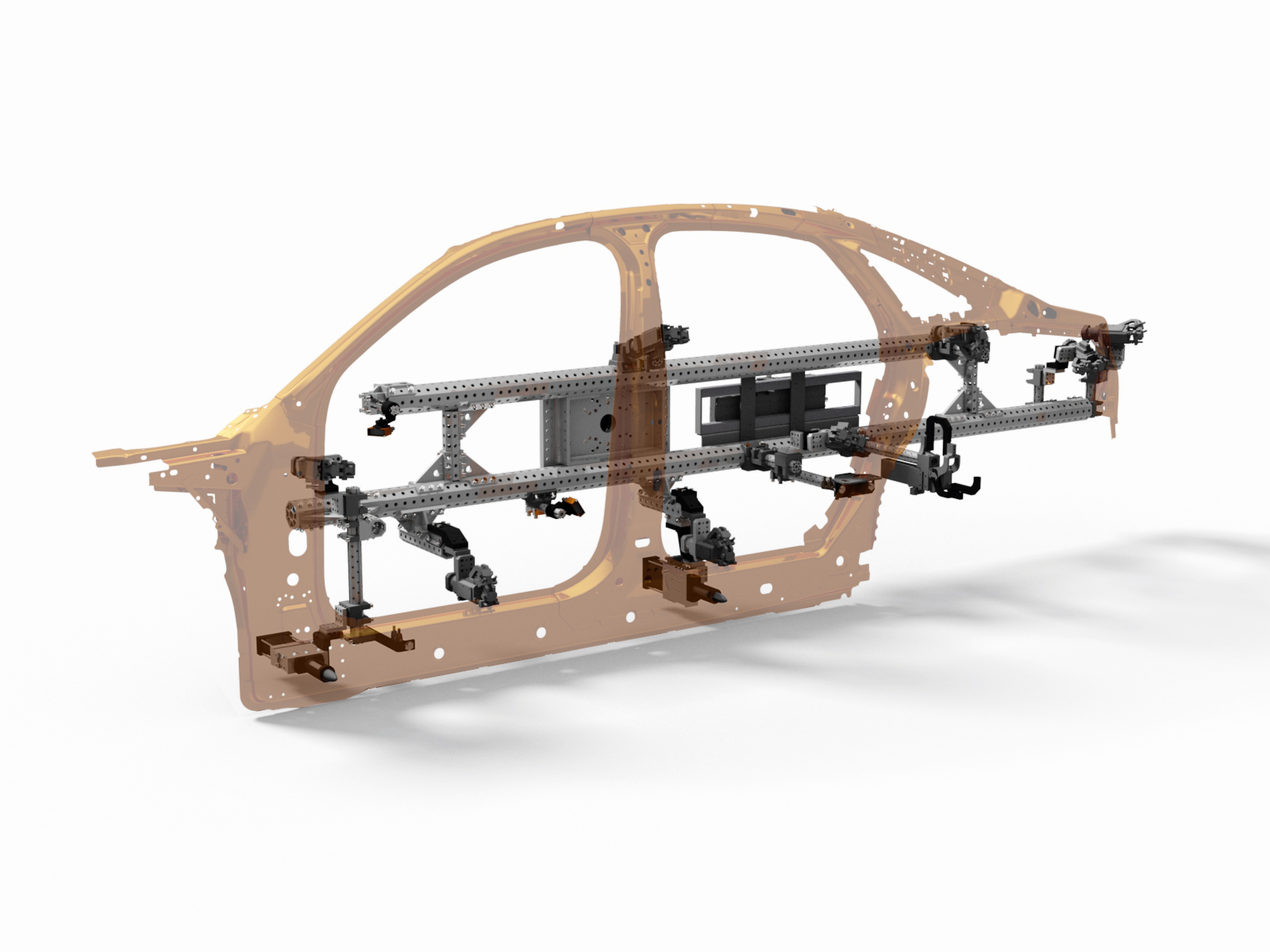}
\caption{Left: Metal piece moved by the arm of the robot. Right: Metal piece as a part of a car.}
\label{fig:metal-piece}
\end{figure}

Due to its complexity, we have understood that the problem cannot be solved in limited time such as five days of the ESGI meeting, but in this time, a conceptual mathematical framework has been laid out, which can be later adapted and solved. Therefore, during the ESGI days, we have restricted the objectives of the problem to the following tasks:

\begin{itemize}

\item Perform a literature search in this area, including a search of numerical codes to solve these kinds of problems.

\item Write the equations of motion of the body-fluid system, in the most general case possible, by determining the stresses and deformations suffered by a body due to the forces of mass and friction acting on it when being moved by another.
    
\item Together with the possible use of commercial, discuss numerical algorithms to solve these equations.

\item Define a very simple case of the problem.  This has been the movement of a longitudinal piece, held at one end, to a body that can move in a direction perpendicular to that of the force of gravity, and can rotate the piece in the plane defined by the direction of motion of the body and that of the force of gravity.
\end{itemize}

The results of the first point will be described in Section \ref{sec:ConcFutureWork} below, together with conclusions about possible future work, and the next three points will be described in Sections \ref{sec:Eqs}, \ref{sec:NumSim}, and \ref{sec:SimpleModelProblem}.

We want to point out that we have devoted our efforts to the so-called {\em direct} problem, rather that the more difficult, but perhaps more important in practice, {\em inverse} problem. Let us explain the difference: In the {\em direct} problem, one assumes that the trajectory of the arm and its rotations are known and given, and the problem is to calculate deformations and stresses of the piece. In the {\em inverse} problem, the goal is to determine trajectories and rotations such that the induced deformations and stresses satisfy the restriction requirements given by the nature of the pieces, to avoid permanent deformations. Even if one considers that only the inverse problem is of practical interest, it is clear that the solution of the inverse problem requires the solution of many direct problems, to achieve the optimal candidate.
\section{The equations}
\label{sec:Eqs}
Before explaining the equations and boundary conditions specific to plate theory, we provide a brief introduction to the governing equations in solid mechanics. This explanation is a condensed version of that found in \cite{Howell2009}, which gives a much more rigorous introduction to the field.
\paragraph{Introduction to solid mechanics}
Suppose that any point inside an elastic object ${\cal P} \subset \mathbb{R}^3$ has an initial \textit{Lagrangian} coordinate $\vec{X}\in {\cal P}$, which is fixed in the material. At time $t>0$, this coordinate changes with elastic deformations of the object, into an \textit{Eulerian} coordinate $\vec{x}(\vec{X},t)$, where the coordinate $\vec{x}$ remains fixed in space. From these, we can construct the \textit{displacement field} $\vec{u}(\vec{x},t):=\vec{x}(\vec{X},t)-\vec{X}$, which in part is found by formulating an equation relating body and surface forces, and using conservation of momentum to obtain a partial differential equation known as the \textit{Navier equation} in which $\vec{u}(\vec{x},t)$ is the dependent variable.

Consider an arbitrary volume $V$ contained within the elastic material. This small volume experiences body forces (e.g. due to gravity) and surface forces. In the case where the density $\rho$ is uniform throughout the material, one can write down expressions for both of these effects, as well as the rate of change of momentum (where momentum is the mass of an object multiplied by its velocity). In this case, the $i$th component of the displacement field, $u_i$, satisfies
\begin{equation*}
\underset{\text{rate of change of momentum}}{\underbrace{\dfrac{{\rm d}}{{\rm d}t}\iiint_{V}\dfrac{\partial u_{i}}{\partial t}\rho{\rm d}{\cal P}}}=\underset{\text{total body force}}{\underbrace{\iiint_{V}g_{i}\rho{\rm d}{\cal P}}}+\underset{\text{total surface force}}{\underbrace{\iint_{\partial V}\sigma_{ij}n_{j}{\rm d}S}},\label{eq:momentum}
\end{equation*}
where $g_i$ is the $i$-th component of the body force, and $\sigma_{ij}$ is the (symmetric) stress tensor, which tells us the $i$th component of the force per unit area acting at a point on the surface with outward normal $n_j$. By taking the time derivative inside the first term (using the fact that the density is constant) and applying the divergence theorem on the final term, we can use the fact that $V$ is arbitrary and the assumption that each integrand is continuous to obtain \textit{Cauchy's momentum equation} 
\begin{equation}
\rho\dfrac{\partial^{2}u_{i}}{\partial t^{2}}=\rho g_{i}+\dfrac{\partial\sigma_{ij}}{\partial x_{j}}.\label{eq:cauchy}
\end{equation}
To be able to determine the displacement field relies on us knowing the stress tensor $\sigma_{ij}$. Thankfully, there is a constitutive relation, known as \textit{Hooke's law}, which postulates a linear relationship between the stress and another quantity called the \textit{strain}, a dimensionless quantity signifying the elastic extension of an object relative to its original state. To identify the exact form of the strain tensor, we can consider two particles with positions $\vec{X}$ and $\vec{X} + \delta\vec{X}$, displaced to $\vec{x}=\vec{X}+\vec{u}(\vec{X},t)$ and $\vec{x}+\delta \vec{x} = \vec{X} + \delta\vec{X} +  \vec{u}(\vec{X} + \delta\vec{X},t)$. Taylor's theorem can be used to show that (upon neglecting quadratic terms)
\begin{equation*}
\left|\delta\vec{x}\right|^{2}=\left|\delta\vec{X}+(\delta\vec{X}\cdot{\nabla}_{\vec{X}})\vec{u}(\vec{X},t)\right|^{2},
\label{eq:strain1}
\end{equation*}
and therefore that
\begin{equation*}
\left|\delta\vec{x}\right|^{2}-\left|\delta\vec{X}\right|^{2}=2\varepsilon_{ij}\delta X_{i}\delta X_{j},
\label{eq:strain2}
\end{equation*}
where the strain tensor $\varepsilon_{ij}$ is given by
\begin{equation}
\varepsilon_{ij}=\dfrac{1}{2}\left(\dfrac{\partial u_{i}}{\partial X_{j}}+\dfrac{\partial u_{j}}{\partial X_{i}}+\dfrac{\partial u_{k}}{\partial X_{i}}\dfrac{\partial u_{k}}{\partial X_{j}}\right)\sim\dfrac{1}{2}\left(\dfrac{\partial u_{i}}{\partial X_{j}}+\dfrac{\partial u_{j}}{\partial X_{i}}\right).\label{eq:strain3}
\end{equation}
If the stress and strain are scalars, then Hooke's law simply states that $\sigma = E \varepsilon$, where $E$ is a constant known as the \textit{Young modulus}. However, in this context, $\sigma_{ij}$ and $\varepsilon_{ij}$ are both rank-2 tensors, so a linear relation between them must involve a rank-4 tensor, so that 
\begin{equation*}
\sigma_{ij}=C_{ijkl}\varepsilon_{kl},\qquad i,j,k,l=1,2,3.
\label{eq:Hooke}
\end{equation*}
Here, $C_{ijkl}$ has 81 entries in total. However, by considering the stress acting at the surface of the volume $V$ (by considering a pillbox-type argument), one can show that the rank-2 tensors are symmetric (i.e. $\sigma_{ij} = \sigma_{ji}$ and $\varepsilon_{ij} = \varepsilon_{ji}$). Under further modelling assumptions, such as homogeneity and isotropy of the material, one can derive a stress-strain relation of the form
\begin{equation}
\sigma_{ij}=\lambda\varepsilon_{kk}\delta_{ij}+2\mu\varepsilon_{ij}\label{eq:Lam=0000E9-stress},
\end{equation}
where $\lambda$ is the bulk modulus and $\mu$ is the shear modulus of the material. Both of these constants are collectively called the \textit{Lam\'{e} constants} and they are material properties which indicate a material's tendency to withstand deformation. Upon substituting \eqref{eq:Lam=0000E9-stress} and \eqref{eq:strain3} into \eqref{eq:cauchy}, one obtains the \textit{Navier equation}, written in vector form as
\begin{equation}
\rho\dfrac{\partial^{2}\vec{u}}{\partial t^{2}}=\rho\vec{g}+(\lambda+\mu){\nabla}({\nabla}\cdot\vec{u})+\mu\nabla^{2}\vec{u}.\label{eq:navier}
\end{equation}
By imposing suitable boundary conditions, one can solve equation \eqref{eq:navier} for the displacement field $\vec{u}(\vec{X},t)$. The usual techniques of applied mathematics, such as asymptotic analysis and numerical methods, can be used to solve this equation, although numerical methods will be much more appropriate for arbitrary geometries (such as those from CAD files provided by component manufacturers).

\paragraph{Classical plate theory}
A considerable simplification in our situation is that many components moved around by a robot are three-dimensional but are thin; for example, a car door. Although many people think of this type of component as strong, namely due to being made of metal, thin metal sheets are prone to elastic displacements which could exceed an elastic limit, deforming plastically and then being unsuitable for use in any further manufacturing (and thus needing to be recycled). Thankfully, a theory has been developed to tackle the case where the component is large and thin. This is called \textit{Kirschoff-Love theory} (or \textit{classical plate theory)}, where in-plane displacements are disregarded, and that displacements normal to the plane are considered relative to a \textit{mid-plane}: a surface equidistant between the top and bottom face of the plane. This idea is illustrated in Figure \ref{fig:plate}, although we emphasise that the plate need not be uniform everywhere since the coordinates are local to the plate.
\begin{figure}
\centering
\includegraphics[width=10cm]{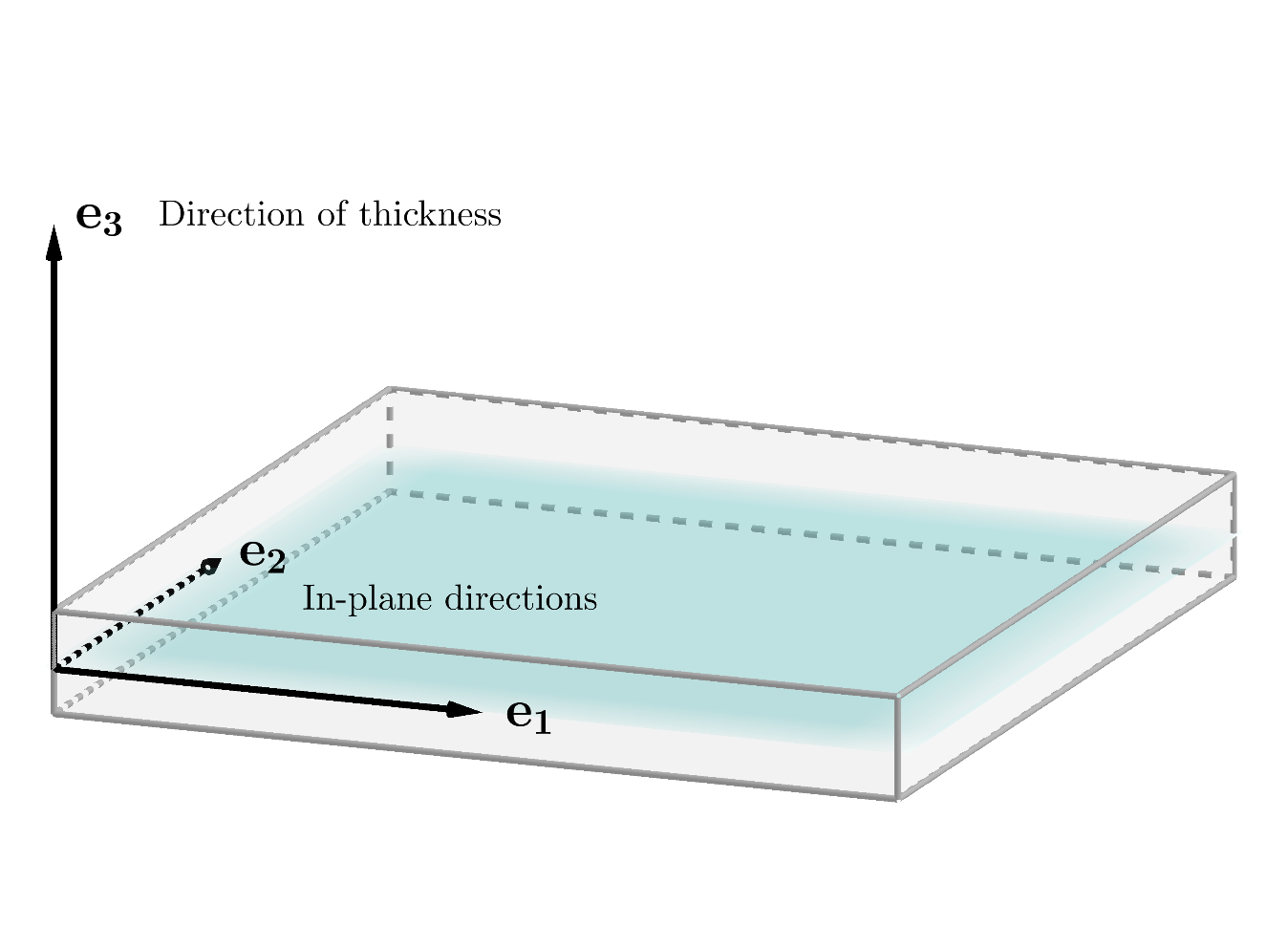}
\caption{A simple illustration of the coordinate system used in Kirchoff-Love plate theory for a (flat) plate.}
\label{fig:plate}
\end{figure}
We guide the reader through the basics of this theory; and refer them to \cite{Reddy2007} for a much more detailed explanation. In the specific case shown in Figure \ref{fig:plate}, an educated guess for the displacement field is given. This is of the form \begin{equation*}
u_{i}(x,y,z,t)=\sum_{j}(z)^{j}\phi_{i}^{(j)}(x,y,t)\label{eq:kirschoff1},
\end{equation*}
where $\phi^{(j)}$ are functions chosen so that the \textit{principle of virtual displacements} is satisfied and $z$ is the coordinate in the direction of thickness. A \textit{virtual displacement} of an elastic system is an infinitesimal displacement which can one might suspect to arise based on the current configuration of forces and if it was perturbed slightly. For example, if an elastic beam of length $L$ is fixed at a wall at $x=0$ and being pulled away by a force $F$ being applied at its free end, then the beam can be thought of as having a ``virtual" displacement of $\delta u(L)$ at its free end arising from a ``virtual" force $\delta F$. The idea of the displacements being ``virtual" is so-called because they are fictional and are unrelated to the displacements and actual loads on the system. 

\paragraph{Virtual work and virtual displacement}
The idea of ``virtual work" arises in two possible ways. We firstly recall that the work done by a force is the product of its projection (dot product) onto its direction of displacement, and its magnitude (or, put simply, ``work is force times distance"). Virtual work can be considered to arise either from (a) an \textit{actual} force moving through a \textit{virtual displacement}, or (b) a \textit{virtual force} moving through an \textit{actual} displacement. If a lowercase delta before a variable denotes an infinitesimal virtual change in that variable, then the virtual work is given by
\begin{equation*}
\delta W=\iiint_{{\cal P}}\vec{F}\cdot\delta\vec{u}{\rm d}{\cal P}.\label{eq:virtual-work}
\end{equation*}
Virtual work can be classified as either \textit{external} or \textit{internal}. Forces applied by external sources will do work when moving through virtual displacements, and to this effect one can write down the \textit{total external virtual work}. When an elastic body is subjected to body forces of $\vec{f}$ per unit volume and surface tractions $\vec{T}$ per unit surface area (where $\Gamma_\sigma$ is the subset of $\partial {\cal P}$ on which stresses are specified), upon moving through virtual displacements $\delta{\vec u}$ it will do a total amount of virtual work $\delta V$ due to applied forces. This is given by
\begin{equation}
 \delta V=-\left(\iiint_{{\cal P}}\vec{f}\cdot\delta\vec{u}{\rm d}{\cal P}+\iint_{\Gamma_{\sigma}}\vec{T}\cdot\delta\vec{u}{\rm d}S\right).\label{eq:external-virtual-work}
\end{equation}
A deforming elastic object will also undergo internal stresses and then the work can be obtained in terms of the total strain inside it rather than due to specific vector forces. In simple terms, the strain of an elastic body is the ratio of the extension of an elastic body to its original size, and is therefore a dimensionless quantity, and since it involves derivatives of displacement, virtual strains $\delta \varepsilon_{ij}$ can be expressed in terms of virtual displacements $\delta u$. By considering work done by normal stresses and shear stresses through virtual displacements, one can show that the total internal work $\delta U$ is given by
\begin{equation}
 \delta U=\sum_{i=1}^{3} \sum_{j=1}^{3}\iiint_{{\cal P}}\sigma_{ij}\delta\varepsilon_{ij}{\rm d}{\cal P}\text{ where }\delta\varepsilon_{ij}=\dfrac{1}{2}\left(\dfrac{\partial\delta u_{i}}{\partial x_{j}}+\dfrac{\partial\delta u_{j}}{\partial x_{i}}\right).\label{eq:internal-virtual-work}
\end{equation}
At this point, we can concisely state the principle of virtual displacements, i.e. that a continuous body in equilibrium will have a total virtual work by actual forces through virtual displacements of zero. Put simply, $ \delta V +  \delta U = 0$; as each of these terms are integrals, the problem then becomes about minimising these integrals (through a \textit{weak formulation}) rather than solving for an unknown function with a differential equation (through a \textit{strong formulation}). More specifically of interest to us, the dynamical version of the principle of virtual displacements is
\begin{equation}
\int_{0}^{T}\left(\delta U+\delta V-\delta K\right){\rm d}t=0\label{eq:PVD-dynamic},
\end{equation}
where $\delta K$ is the virtual kinetic energy of the system.
This method of deriving governing equations for an elastic body is an attractive alternative to equation \eqref{eq:navier} because (a) it does not need to rely on the use of constitutive laws upfront (e.g. Hooke's law), and (b) the elasticity problem can be posed in the form of a variational problem, thereby making it amenable to both analytical treatments (e.g. calculus of variations, Euler-Lagrange equations, Hamilton's principle) and contemporary numerical methods such as finite element methods (FEM). How one might approach this problem using FEM is to be discussed later.

\paragraph{Deriving the governing equation}

We are interested in \textit{Kirchoff's hypothesis} which provides the elastic displacement field \textit{a priori} as
\begin{align*}
u(x,y,z,t) & =u_{0}(x,y,t)-z\dfrac{\partial w_{0}}{\partial x},\\
v(x,y,z,t) & =v_{0}(x,y,t)-z\dfrac{\partial w_{0}}{\partial y},\\
w(x,y,z,t) & =w_{0}(x,y,t),
\end{align*}
where $u_0$, $v_0$ and $w_0$ are displacements from the mid-plane in the $x$, $y$ and $z$-directions respectively. In particular, we can set $u_0\equiv 0$ and $v_0 \equiv 0$, due to neglecting in-plane displacements. Namely, we are interested in the domain $[-w,w]\times \cal{F}$, where $\cal{F}$ represents the cross-section of the plate at $z=0$. Substituting the above form of the displacement field into the linearised strains \eqref{eq:strain3}, neglecting in-plane displacements, it is straightforward to show that the strain tensor is given by
\begin{equation}
\label{eq:strainK-11}
    \begin{aligned}
\epsilon_{11} & =\dfrac{1}{2}\left(\dfrac{\partial w_{0}}{\partial x}\right)^{2}-z\dfrac{\partial^{2}w_{0}}{\partial x^{2}}\\
\epsilon_{22} & =\dfrac{1}{2}\left(\dfrac{\partial w_{0}}{\partial y}\right)^{2}-z\dfrac{\partial^{2}w_{0}}{\partial y^{2}}\\
\epsilon_{12} & =\dfrac{1}{2}\left(\dfrac{\partial w_{0}}{\partial x}\dfrac{\partial w_{0}}{\partial y}-2z\dfrac{\partial^{2}w_{0}}{\partial x\partial y}\right)\\
\epsilon_{13} & =\dfrac{1}{2}\left(-\dfrac{\partial w_{0}}{\partial x}+\dfrac{\partial w_{0}}{\partial x}\right)=0\\
\epsilon_{23} & =\dfrac{1}{2}\left(-\dfrac{\partial w_{0}}{\partial y}+\dfrac{\partial w_{0}}{\partial y}\right)=0\\
\epsilon_{33} & =0,    
    \end{aligned}
\end{equation}
where we have given six entries instead of nine, owing to the fact that the strain tensor is symmetric (so that $\epsilon_{ij}=\epsilon_{ji}$). This puts us in a position to calculate the internal virtual work $ \delta U$, given by
\begin{align*}
 \delta U & =\int_{-w}^{w}\int_{{\cal F}}\left(\sigma_{11}\delta\epsilon_{11}+2\sigma_{12}\delta\epsilon_{12}+\sigma_{22}\delta\epsilon_{22}\right){\rm d}{\cal F}{\rm d}z\nonumber \\
 & =-\int_{-w}^{w}\int_{{\cal F}}\left(\sigma_{11}z\dfrac{\partial^{2}\delta w_{0}}{\partial x^{2}}+2\sigma_{12}z\dfrac{\partial^{2}\delta w_{0}}{\partial x\partial y}+\sigma_{22}z\dfrac{\partial^{2}\delta w_{0}}{\partial y^{2}}\right){\rm d}{\cal F}{\rm d}z + \text{h.o.t.} \\
 & =-\int_{{\cal F}}\left(M_{11}\dfrac{\partial^{2}\delta w_{0}}{\partial x^{2}}+2M_{12}\dfrac{\partial^{2}\delta w_{0}}{\partial x\partial y}+M_{22}\dfrac{\partial^{2}\delta w_{0}}{\partial y^{2}}\right){\rm d}{\cal F}+\text{h.o.t.}\label{eq:plateInternalStrain}
\end{align*}
where ``h.o.t." is an abbreviation for ``higher-order terms"; terms which are quadratic in the partial derivatives of $\delta w_0$ and can therefore be neglected (although this is no longer appropriate if von Karman strains are considered).
We also have
\begin{equation}
M_{ij}=\int_{-w}^{w}\sigma_{ij}z{\rm d}z\label{eq:M_ij},
\end{equation}
are the \textit{stress moment resultants}. We emphasise that in general, there will also be \textit{stress resultants}
\begin{equation*}
N_{ij}=\int_{-w}^{w}\sigma_{ij}{\rm d}z\label{eq:N_ij},
\end{equation*}
but neglecting the terms which are quadratic in the derivatives of $w_0$ leads to them being absent from the expression for the internal virtual work.

Meanwhile, the external virtual work is given by
\begin{align}
 \delta V & =-\int_{-w}^{w}\int_{{\cal F}}\vec{f}\cdot\delta\vec{u}{\rm d}{\cal F}{\rm d}z-\int_{-w}^{w}\int_{\partial{\cal F}}\vec{T}\cdot\delta\vec{u}{\rm d}S{\rm d}z\label{eq:external-virtual-work0}\\
 &=-\int_{-w}^{w}\int_{{\cal F}}\left(q-kw_{0}\right)\delta w_{0}{\rm d}x{\rm d}y\label{eq:external-virtual-work}.
\end{align}

The second term in equation \eqref{eq:external-virtual-work0} disappears because the traction force, $T_{i}=\sigma_{ij}n_{j}$
where $\vec{n}$ is the normal to $\partial{\cal F}$ (a closed curve in the $(x,y)$-plane), is perpendicular to the $z$-direction in which the displacement is solely assumed to take place. In equation \eqref{eq:external-virtual-work}, $q(x,y)$ is the net load on $\cal{F}$, and the $-kw_0$ contribution arises from Hooke's law, where $k$ is the stiffness constant of the plate material. 
The internal kinetic energy $\delta K$ is given by
\begin{align}
\delta K & =\int_{{\cal P}}\int_{-w}^{w}\rho\left(\dot{u}\delta\dot{u}+\dot{w}\delta\dot{w}+\dot{w}\delta\dot{w}\right){\rm d}z{\rm d}x{\rm d}y\nonumber \\
 & =-I_{0}\int_{{\cal P}}\left(\dot{u}_{0}\delta\dot{u}_{0}+\dot{w}_{0}\delta\dot{w}_{0}+\dot{w}_{0}\delta\dot{w}_{0}\right){\rm d}x{\rm d}y-I_{2}\int_{{\cal P}}\left(\dfrac{\partial\dot{w}_{0}}{\partial x}\dfrac{\partial\delta\dot{w}_{0}}{\partial x}+\dfrac{\partial\dot{w}_{0}}{\partial y}\dfrac{\partial\delta\dot{w}_{0}}{\partial y}\right){\rm d}x{\rm d}y\label{eq:internal-kinetic-energy}
\end{align}
where the moments of inertia are given by
\begin{equation*}
I_{k}=\int_{-w}^{w}z^{k}\rho{\rm d}z\text{ where }k=0,1,2.\label{eq:moment-of-inertia}
\end{equation*}
We do not present the full derivation of the governing equations here due to the amount of algebra involved, but briefly describe how it is done (see pages 103-105 of \cite{Reddy2007} for a full derivation). By substituting results \eqref{eq:internal-virtual-work}, \eqref{eq:external-virtual-work} and \eqref{eq:internal-kinetic-energy} into \eqref{eq:PVD-dynamic}, making use of the virtual strains, employing techniques from the calculus of variations and setting the coefficient of $\delta w_0$ to zero, one can show that
\begin{equation*}
\dfrac{\partial^{2}M_{11}}{\partial x^{2}}+2\dfrac{\partial^{2}M_{12}}{\partial x\partial y}+\dfrac{\partial^{2}M_{22}}{\partial y^{2}}-kw_{0}+q=I_{0}\dfrac{\partial^{2}w_{0}}{\partial t^{2}}-I_{2}\dfrac{\partial^{2}}{\partial t^{2}}\left(\dfrac{\partial^{2}w_{0}}{\partial x^{2}}+\dfrac{\partial^{2}w_{0}}{\partial y^{2}}\right).\label{eq:moment-EOM}
\end{equation*}
Finally, it is desirable to express this governing equation in terms of displacements rather than moments. For a homogeneous, isotropic plate, a constitutive law tells us that
\begin{equation}
    	\left(\begin{matrix}\sigma_{11}\\
	\sigma_{22}\\
	\sigma_{12}
	\end{matrix}\right)=\dfrac{E}{1-\nu^{2}}\left(\begin{matrix}\begin{matrix}1 & \nu & 0\\
	\nu & 1 & 0\\
	0 & 0 & 1-\nu
	\end{matrix}\end{matrix}\right)\left(\begin{matrix}\epsilon_{11}\\
	\epsilon_{22}\\
	\epsilon_{12}
	\end{matrix}\right)\label{eq:stress-strain},
\end{equation}
where $E$ is the Young modulus of the material, and $\nu$ is the Poisson ratio. Combining the constitutive law \eqref{eq:stress-strain} with \eqref{eq:M_ij} and the strain-displacement relations \eqref{eq:strainK-11}, one obtains 
	\begin{equation}
	D \nabla^2 \nabla^2 w_0 = - q(x,y,t) - 2 \rho h \frac{\partial^2 w_0}{\partial t^2},\label{eq:Kirschoff-Love}
	\end{equation}
	where the \textit{bending stiffness}, $D$, is given by
	\begin{equation}
D=\dfrac{2h^{3}E}{3(1-\nu^{2})},\label{eq:bending-stiffness}
\end{equation}
which will be loosely referred to as the \textit{``Kirchoff-Love equation"}.

\paragraph{Initial and boundary conditions}
Along with the equation of motion \eqref{eq:Kirschoff-Love}, several boundary conditions based on likely physics in a factory setting need to be imposed. Let $\mathcal{C}\subset\partial \mathcal{F}$ be the part of the large face boundary which is clamped by the robot, then the plate
	\begin{enumerate}
		\item is \textit{initially undeformed}, i.e.,
		$$
		w_0(x,y,t=0)=0,
		$$

		\item is \textit{initially stationary}, i.e.,
		$$
		\frac{\partial w_0}{\partial t}(x,y,t=0)=0,
		$$
	
		\item \textit{does not deform} where it is clamped, i.e.,
		$$
		w_0(x,y,t)=0, \ x,y\in \mathcal{C};
		$$

		\item and \textit{no bending moments or loads} at the free boundary:
		$$
		\nabla^2 w_0(x,y,t) = 0, \
		\frac{\partial}{\partial n} (\nabla^2 w_0(x,y,t)) = 0, \ (x,y) \in\partial \mathcal{F} \backslash \mathcal{C}
		$$
	\end{enumerate}

\paragraph{Changing the frame}
As it stands, it suffices to solve equation \eqref{eq:Kirschoff-Love} for the transverse elastic displacements of the plate, provided that the plate is not being moved externally. Considering how the plate is being moved around by a robot, this is not satisfactory; if the plate is being moved in a non-inertial frame, it will experience ``fictitious" forces, namely the Euler, Coriolis and centrifugal forces. These forces are so-called because they arise from a change of frame rather than a physical mechanism. They are mathematical in nature and should therefore be incorporated within the external load $q(x,y,t)$ as it appears in equation \eqref{eq:Kirschoff-Love}. The mathematical techniques for changing frames are standard and can be found in classical mechanics texts such as \cite{goldstein2002classical}.

\begin{figure}
\centering
\includegraphics[width=15cm]{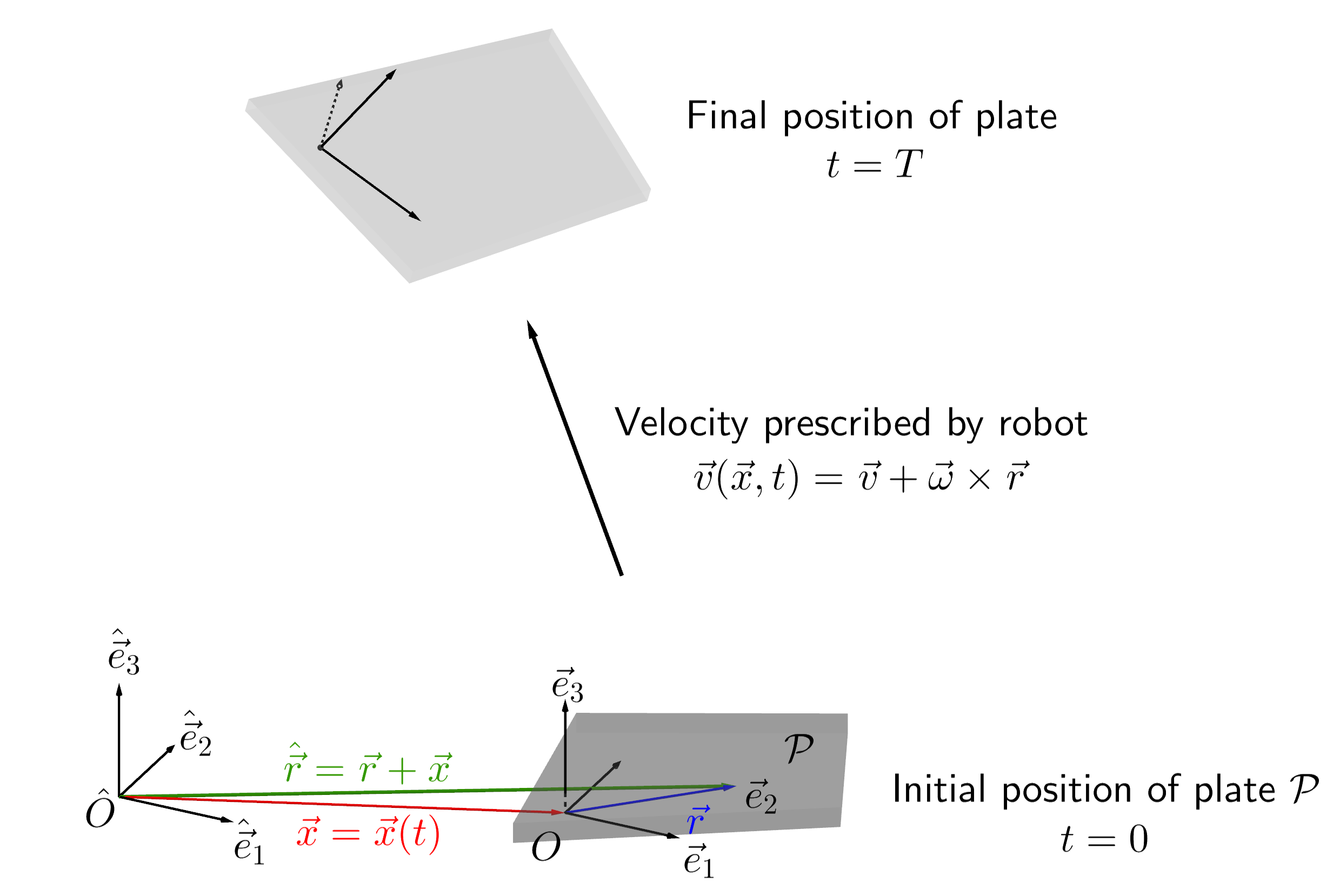}
\caption{An illustration of plate motion in two frames of reference.}
\label{fig:moving-plate}
\end{figure}

We use $\hat{\cal{S}}$ to denote an \textit{inertial} frame which remains fixed over time, with origin $\hat{O}$ which is fixed in space, and orthonormal basis $\{\hat{{\vec e}}_1,\hat{{\vec e}}_2,\hat{{\vec e}}_3\}$. We use $\cal{S}$ to denote the \textit{non-inertial} frame, which is fixed local to the plate and whose origin $O$ moves with angular velocity $\vec{\omega}$ relative to $\hat{O}$. The relationship between the two coordinate systems is illustrated in Figure \ref{fig:moving-plate}.

F.EE have complete control over the rotational movement of the robotic arm, as well as translational motion along a straight line. In practice, the rotational motion of the robot motion may be prescribed by a \textit{rotation matrix} ${\cal R}_{ij}(t)$ which acts as a time-dependent linear transformation from $\hat{\cal{S}}$ to $\cal{S}$ and may be expressed in terms of Euler angles. If $\hat{{\rm D}}$ and ${\rm D}$ denote time derivatives in the inertial and non-inertial frames respectively, and $\vec{r}$ 
is the position vector of a point in $\cal{S}$ fixed on the plate relative to $O$, then the Coriolis formula is given by 
\begin{align*}
\hat{{\rm D}}\vec{r} & ={\rm D}\vec{r}+\vec{\omega}\times\vec{r},\label{eq:Coriolis}
\end{align*}
where $\vec{\omega}$ is the \textit{angular velocity} vector whose entries satisfy
\begin{equation}
{\cal R}_{ij}\dot{{\cal R}}_{ji}=\sum_{k=1}^{3}\epsilon_{ijk}\omega_{k}.\label{eq:Omega-omega}
\end{equation}
Applying the Coriolis formula twice can be used to obtain an expression for the acceleration $\hat{\vec{a}}$ in the inertial frame in terms of the acceleration $\vec{a}$ in the non-inertial frames:
\begin{equation}
\hat{\vec{a}}=\vec{a}+({\rm D}\vec{\omega})\times\vec{r}+2\vec{\omega}\times{\rm D}\vec{r}+\vec{\omega}\times(\vec{\omega}\times\vec{r})+\vec{A}, \label{eq:Coriolis-acceleration}
\end{equation}
where $\vec{A}=\hat{{\rm D}}^2 \vec{x}$ is the acceleration of $O$ relative to $\hat{\cal{S}}$.

This is useful because it can be used in principle to re-express the Kirchoff-Love equation \eqref{eq:Kirschoff-Love} in a non-inertial frame local to the plate, as the acceleration arises naturally there. The acceleration formula \eqref{eq:Coriolis-acceleration} can then be used to to re-express Newton's second law in $\cal{S}$ rather than $\hat{\cal{S}}$:
\begin{align}
m\hat{\vec{a}} & =\vec{F}\text{ in }\hat{{\cal S}}\implies m\vec{a}=\vec{F}-\underset{\text{``Euler force"}}{\underbrace{m\left({\rm D}\vec{\omega}\right)\times\vec{r}}}-\underset{\text{``Coriolis force"}}{\underbrace{2m\vec{\omega}\times{\rm D}\vec{r}}}-\underset{\text{``Centrifugal force"}}{\underbrace{m\vec{\omega}\times\left(\vec{\omega}\times\vec{r}\right)}}+m\vec{A}\text{ in }{\cal S}.\label{eq:N2}
\end{align}
Since we are only concerned with elastic oscillations of the plate in the $\vec{e}_3$-direction, it suffices to take the dot product of equation \eqref{eq:N2} with $\vec{e}_3$ in order to obtain the equivalent of the Kirschoff-Love equation \eqref{eq:Kirschoff-Love}. We do not write down the equation in full, since it depends on the specific rotations being applied. However, we simply state that the term involving the second derivative of $w_0$ with respect to time corresponds to $\textbf{a}\cdot \textbf{e}_3$ and that the external loads $q(x,y,t)$ correspond to $\vec{F}$.

It is not quite enough to simply apply a change of frame to the Kirschoff-Love equation to take all physical considerations into account. There is also the weight of the plate, the inertia, the internal elastic forces inside the plate, and also the air drag. However, in a factory setting, the effects of these terms would be questionable, and may make the equations significantly more coupled without adding much insight. This is particularly the case for the air drag because a complete description would rely on coupling the elastic displacement equations discussed hitherto with the Navier-Stokes equations of fluid dynamics. A simplified starting point could be to introduce a simple drag law, where the drag force on an object in a fluid is proportional to the square of the speed at which object is being passed by the fluid. The exact nature of the physical effects is well beyond the scope of this report, but would be interesting nonetheless. 

Another physical phenomenon to consider is that many elastic materials deform plastically when their displacements pass a limit that is large enough in size. This is often called \textit{yield}, and a hypothesis governing an instance in which a material deforms is known as a \textit{yield criterion}. Yield hypotheses are frequently given in terms of functions of the entries of the stress tensor ${\sigma}$; for example, the \textit{von Mises' stress criterion} \cite{vonMises}
\begin{equation*}
\sigma_{{\rm VM}}=\sqrt{\dfrac{3}{2}(\sigma_{11}^{2}+2\sigma_{12}^{2}+\sigma_{22}^{2})-\dfrac{1}{2}(\sigma_{11}^{2}+\sigma_{22}^{2})}.
\end{equation*}
This is a nonlinear equation on the stresses inside the elastic plate, and therefore the strains and the elastic displacements. The critical stress $\sigma_\text{VM}$, if exceeded by the elastic plate, will result in it failing to revert to its original state. The von Mises' stress criterion could be a useful diagnostic in designing possible paths for the robotic arm to take; since the stress tensor entries are determined from the elastic displacements, this could result in the suggestion of a suitable inverse problem: \textit{given that the stress cannot exceed a certain limit, how should the path of the robotic arm be constructed?}

We also note that, in a factory setting, the time variation in the Kirschoff-Love equation \eqref{eq:Kirschoff-Love} is likely to be relatively slow. This indicates that the time-dependence of $q$ and the time-derivative can be neglected, although one needs to check that this is justified by nondimensionalising and using typical scalings provided by those at F.EE. Upon doing so, the Kirschoff-Love equation reduces to a biharmonic equation $D\nabla^4 w_0 = -q(x,y)$, which can be solved using asymptotic methods and possibly (depending on the form of $q$ and the presence of the non-inertial terms) complex variable methods. A simplifying assumption could be to use the \textit{quasi-static hypothesis}, which takes out the time dependence initially to find a solution, and then reinstates the time-dependence in $q$ once this is done (for example, by seeking an asymptotic  expansion solution to $w_0$ which is steady at leading-order).

\section{The numerical simulation}
\label{sec:NumSim}
\noindent
In this section, we outline the steps towards a possible simulation with a simple
model of a solid shell using Finite Element Analysis, FEA
(See~\cite{ZienkiewiczT2000-Vol2}, for the application of FEA to solid mechanics).

The purposed model consists of a shell of small thickness moving on its
transversal direction (see Figure~\ref{fig:plate}), $z$-axis in the figure, with
constant acceleration, $\vec{a} = (0,0,a_{z})$. We consider non-inertial forces,
the weight of the plate, and friction with the air. 

This model is suitable to be analyzed by means of the FEA, for which the main steps are:
meshing the domain, then introduce the weak formulation to set up the local
equations for the elements, that are coupled to form the \emph{global system}.
The boundary conditions of the original problem are imposed on this system of
linear equations to give a \emph{reduced system} whose solution, the \emph{nodal
solution}, consists of the displacements of the nodes with respect to the
undeformed body. (and other quantities, depending on the element type chosen to
carry out the analysis, see below).   Later, in the post-process, one can compute
other results derived from the nodal solution. Typically, the strains and
stresses and the Von Mises' stress.

\begin{figure}[t]
\centering
   \includegraphics[scale=0.4]{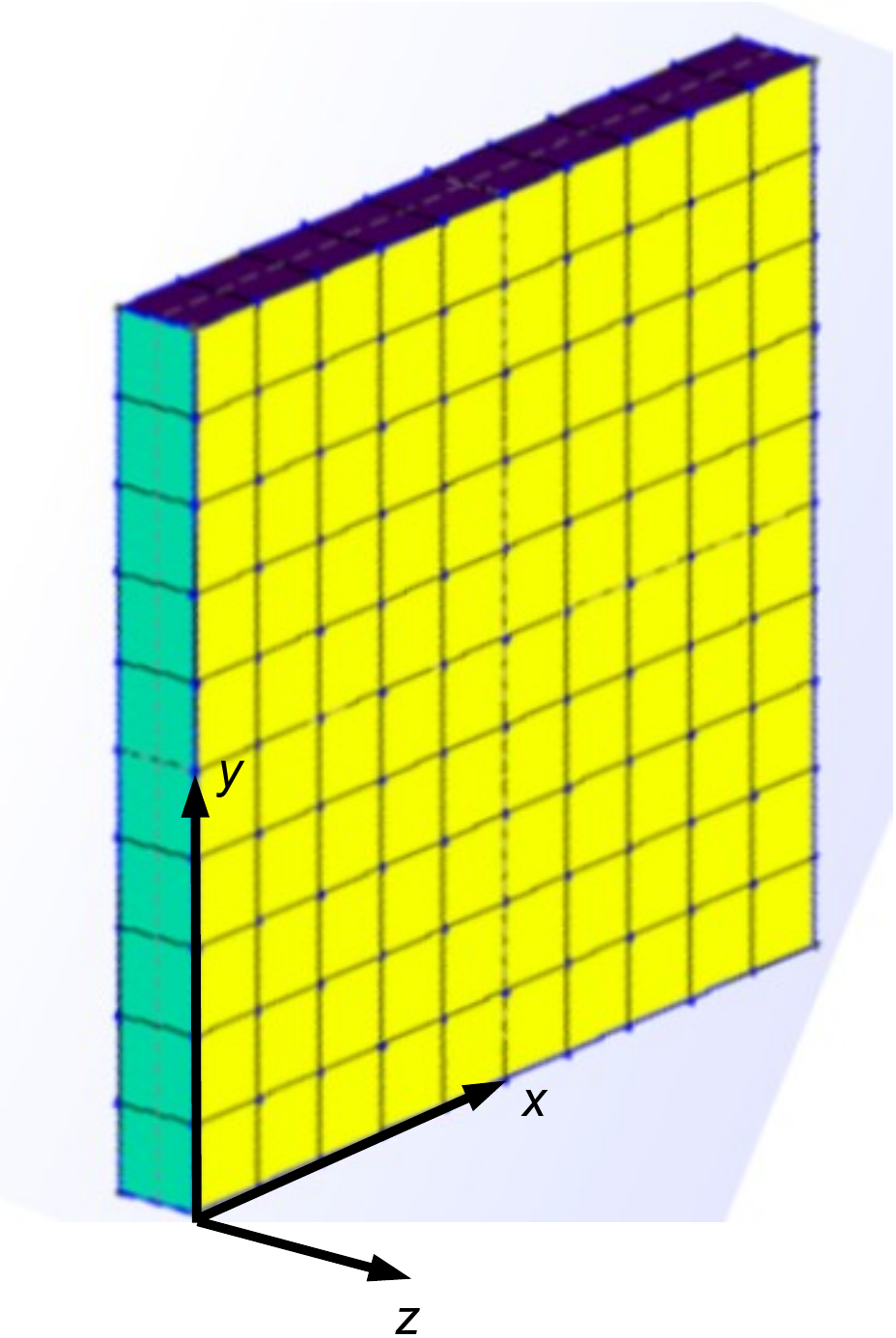}
   \caption{Meshing the 3D domain: element and nodes. Source: 
   \url{https://openfoamwiki.net}\label{fig:plate}}
\end{figure}

\vspace{\baselineskip}
\noindent
\textbf{ Weak formulation and the principle of virtual work.}
The formulation of the weak form of the equations from the principle of
virtual work is presented in many books in solid mechanics. Here, we give a
brief account and point the reader to~\cite[Chapter 1]{Lush2014}, and references
therein. 

So, basically, to find the local equations of the element, ${\cal P}^{e}$, the
underlying idea is that the changes in its internal strain energy are equal to the
virtual work of the external forces acting on it. Therefore,
\begin{equation}\label{eq:vw}
\underbrace{\iiint_{{\cal P}^{e}}
	\delta\varepsilon^{\top}
	\sigma\;\mathrm{d} V}_{\begin{array}{l}
	\text{\small Virtual changes in}\\
	\text{\small the internal strain}\\
	\text{\small energy}
\end{array}} = \underbrace{
		\iiint_{{\cal P}^{e}}
		\delta\ell^{\top} f_{b}\; \mathrm{d}V +
		\iint_{\partial{\cal P}^{e}}
		\delta\ell^{\top} f_{t}\; \mathrm{d}S
+\dots}
_{\begin{array}{l}
	\text{\small Virtual work made by}\\
	\text{\small the external forces}
\end{array}},
\end{equation}
with $\varepsilon$ being the strain, $\sigma$ the stress, $\ell$ the vector
displacement of the nodes, $f_{b}$ the body forces (weight and non-inertial
forces) and $f_{t}$ the surface forces (note that the integral is taken on the
boundary of the element) whereas $\delta\varepsilon$ and $\delta\ell$ denote
the virtual changes on strain and displacements, respectively.

In a linear analysis context, the strain and displacements are related by the strain-displacement matrix, $B$, as
$\varepsilon = B\ell$, so for virtual displacements, $\delta\varepsilon =
B\delta\ell$. On the other hand, the stress-strain law states that $\sigma =
C\varepsilon$. Then, the substitution of these relations in~\eqref{eq:vw} yields

\begin{displaymath}
\delta\ell^{\top}
\equalto{\underbrace{
         \left(\iiint_{{\cal P}^{e}} B^{\top}C B\mathrm{d} V\right)}}%
	{K^{e}}\ell^{e} = 
\delta\ell^{\top}
\equalto{\underbrace{
         \left(\iiint_{{\cal P}^{e}} f_{b}\; \mathrm{d} V +
               \iint_{\partial{\cal P}^{e}}
		f_{t}\; \rmd S
		+\dots\right)}}%
	{R^{e}_{ext}},
\end{displaymath}
and (arbitrary) virtual displacements, $\delta\ell$, cancel out to give,
\begin{equation}
K^{e} \ell^{e} = R^{e}_{ext}, 
\label{eq:weak-form}
\end{equation}
which are the local equations for the element ${\cal P}^{e}$. The integrals are
computed using Gaussian integration (see~\cite[Chapter 5]{Bathe2006}, for an 
account of Gaussian integration in FEA related problems).

\vspace{\baselineskip}
\noindent
\textbf{Nodal solution and post-process.} Note that
$K^{e}$ and $R^{e}_{ext}$ are the stiffness matrix and the force
vector of element ${\cal P}^{e}$, respectively. These local equations must be assembled to form
the global system. Next, the boundary conditions of the continuous model must be
tailored to be set at the (global) nodes of the assembled system. No traction
nor compression is assumed to act at the boundaries, so we consider only
\emph{essential} boundary conditions: the displacement of the nodes grabbed by
the grips are fixed to zero. The outcome is a \emph{reduced} system of linear
equations, where only the displacements of \emph{free} (not fixed) nodes remain
as unknowns. In the end, at the post-process, the strain and the stress at the
nodes are computed from the displacements using the above-mentioned relations,
i.e.,
\begin{displaymath} 
	\varepsilon^{e} = B\ell^{e},\qquad\qquad \sigma^{e} =
C\varepsilon^{e} = C B\ell^{e}; 
\end{displaymath} 
moreover, von Mises' stress $\sigma^{e}_{VM}$ can be derived from the components
of $\sigma^{e}$.  

\vspace{\baselineskip} \noindent 
\textbf{Meshing the 3D domain and MITC9 shell element.} In FEA, it is worth choosing the most suitable element type to mesh the domain.
In the case of a car's bodywork, where the thickness is small in front of the
surface extension, and curves surfaces are present, shell finite elements are
widely used. For a description, see~\cite[Chapter 5]{Bathe2006};
and~\cite[Chapter 11]{Reddy2007}, for a thorough analysis of shell elements and a complete list of references.  In~\cite[Chapter 3]{Lush2014}, IDC
(Isoparametric Degenerate Continuum) and MITC9 (Mixed Interpolation of Tensorial
components with $9$ nodes) shell elements are explained in more detail. In
particular, we propose the use of this last one to analyze the car's bodywork.
The formulation of the MITC9 finite element we comment below is taken from a
more recent work~\cite{Moysidis2019}, where this element is extended to the
elastoplastic analysis of shell structures by adding and hysteresis model. 

We remark that, as pointed in~\cite{Lush2014}, the main characteristic of the
MITC family of finite elements is that, in order to avoid the shear-locking
phenomenon (due to the small thickness, nodes in that direction are very close
and the associated coefficients in the stiffness matrix become very large), it
uses separate interpolation functions for the tangent and transverse shear
strain components. 

\begin{figure}[!b]
\centering
\includegraphics[scale=0.4]{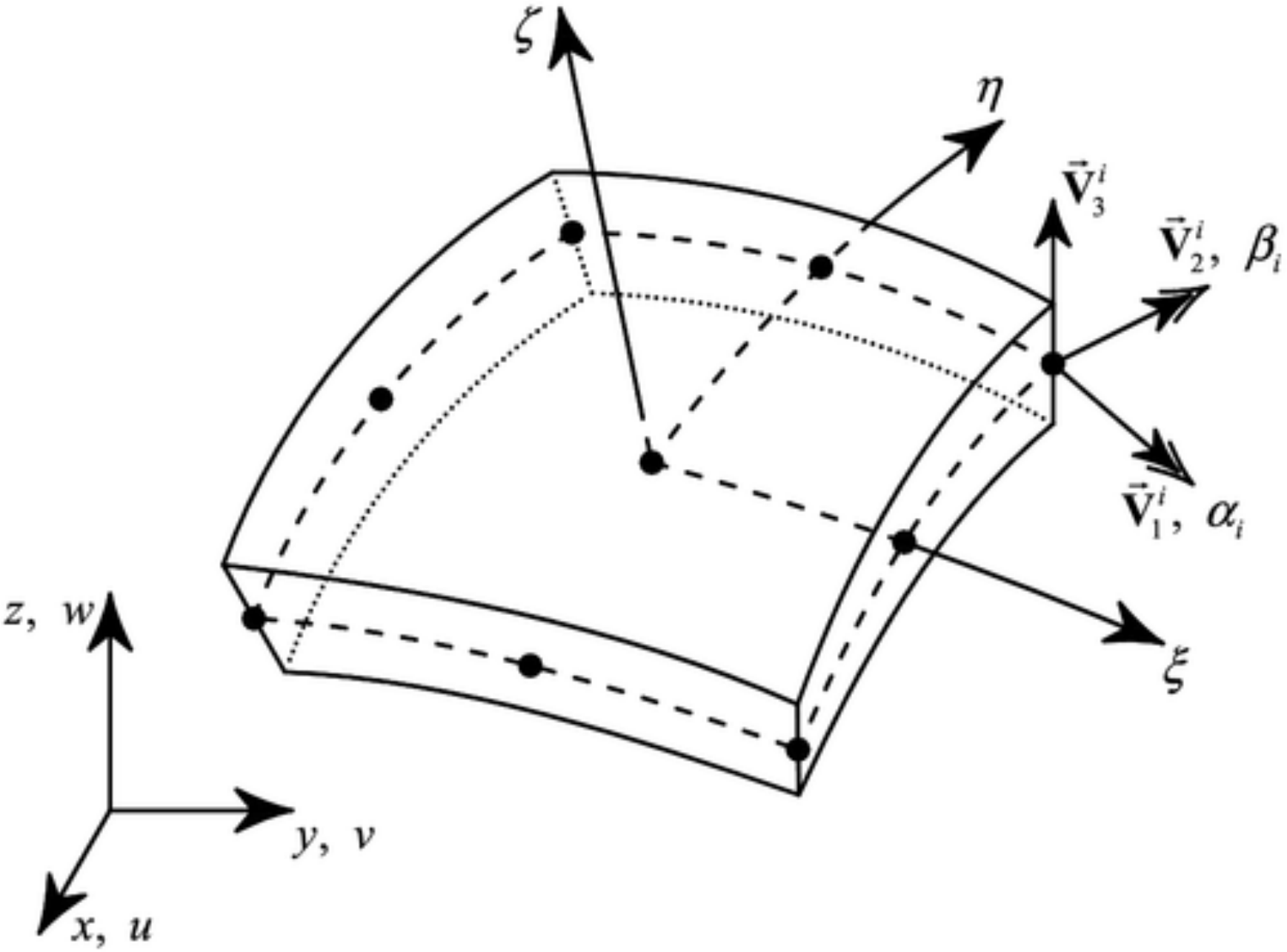}
\caption{MITC9 element. Source: A.N.~Moysidis and V.K.~Koumousis 
(see~\cite{Moysidis2019}).}\label{fig:MITC9}
\end{figure}

The aspect of the MITC9 element is shown in Figure~\ref{fig:MITC9}. The $9$
nodes are placed in the middle surface. The coordinates $\xi$, $\eta$, and
$\zeta$ are local coordinates ``attached'' to the central node; $\xi$, $\eta$
are curvilinear coordinates whereas axis $\zeta$ points in the direction of the
thickness (therefore, it is not exactly normal to the mid-surface).
Furthermore, associated with each node, there are three orthogonal vectors
$V_{1i}$, $V_{2i}$ and $V_{3i}$. For each $i=1,\dots,9$, $V_{3i}$ is the vector
on the line defined by the thickness from the bottom to the top of the element.
If $p_{i}^{\top} = (x_{i},y_{i},z_{i})$, $i=1,\dots,9$, are the Cartesian
coordinates of the nodes, then the Cartesian coordinates of any point of
the element are given in terms of the local element coordinates $\xi$, $\eta$,
$\zeta$, by
\begin{equation}\label{eq:xyz}
	\begin{pmatrix}
	x\\ y\\ z\end{pmatrix} =
	\sum_{i=1}^{9} N_{i}(\xi,\eta)
	\begin{pmatrix}
	x_{i}\\ y_{i}\\ z_{i}\end{pmatrix} +
	\frac{\zeta}{2}\sum_{i=1}^{9} N_{i}(\xi,\eta) V_{3i},\qquad\quad
\end{equation}
here $N_{i}(\xi,\eta)$, $i=1,\dots,9$, are the shape functions
(see~\cite[ Chapter 3]{Lush2014}). To define the
two vectors orthogonal to each $V_{3i}$ one way to proceed is to take
$V_{i1} = \hat{\jmath} \times V_{3i}$, $V_{1i} = V_{3i}\times V_{1i}$, where
$\hat{\jmath}$ is the unit vector in the direction of the $y$-axis. Let
$\hat{v}_{1i}$, $\hat{v}_{2i}$ be the normalized vectors of $V_{1i}$ and
$V_{2i}$, and $\alpha_{i}$, $\beta_{i}$ be rotations of vector $V_{3i}$ about
$\hat{v}_{1i}$ and $\hat{v}_{2i}$, respectively. If in addition, it is assumed
that the angles $\alpha_{i}$, $\beta_{i}$, $i=1,\dots,9$ and the 
strains in the direction normal to the middle surface are small, then the
displacements $u, v, w$ in the Cartesian coordinates $x, y, z$ (see
Figure~\ref{fig:MITC9}) are given by the formula,
\begin{equation}\label{eq:uvw}
\begin{pmatrix}
	u\\ v\\ w\end{pmatrix} =
	\sum_{i=1}^{9} N_{i}(\xi,\eta)
	\begin{pmatrix}
	u_{i}\\ v_{i}\\ w_{i}\end{pmatrix} +
	\frac{\zeta}{2}\sum_{i=1}^{9} t_{i} N_{i}(\xi,\eta)
	\left(-\alpha_{i}\hat{v}_{2i}+\beta_{i}\hat{v}_{1i}\right),
\end{equation}
here $t_{i}$ is the thickness of the element at the $i$-th node for $i=1,\dots,9$.
Thus, the displacements $(u,v,w)$ for any point of the element are determined 
by the displacements $(u_{i},v_{i},w_{i})_{i=1,\dots,9}$ of the nodes at the
mid-surface and rotation angles $(\alpha_{i},\beta_{i})_{i=1,\dots,9}$.

From~(\refeq{eq:uvw}) we can compute the derivatives of the displacements with
respect to the local element coordinates: $\partial_{\xi}u(\xi,\eta,\zeta),
\partial_{\eta} u(\xi,\eta,\zeta), \partial_{\zeta}u(\xi,\eta,\zeta), \dots$
(the same for $v$ and $w$). 

Finally, to write down the strain displacement-matrix $B$ and later compute the
stiffness matrix of the element (see~\cite[Chapter 3]{Lush2014}, and references
therein for explicit formulas of $B$ as well as for the stress-strain matrix
$C$), the derivatives of the displacements with respect to the Cartesian
coordinates $x, y, z$ are required, but these follow at once from the relations
\begin{displaymath}
	\nabla \hat{u}(x,y,z) = 
	\left[\frac{\partial(x,y,z)}{\partial(\xi,\eta,\zeta)}\right]^{-\top}
		\nabla u(\xi,\eta,\zeta),\dots  
		\text{(the same for $\hat{v}$ and $\hat{w}$),}
\end{displaymath}
where $\hat{u}, \hat{v}, \hat{w}$ are the displacements as functions of the
Cartesian coordinates, $\nabla$ is the usual notation for the gradient operator
(actually, $\nabla = (\partial/\partial x,\partial/\partial y,\partial/\partial
z)^{\top}$ at the l.h.s.~and $\nabla =
(\partial/\partial\xi,\partial/\partial\eta,\partial/\partial\zeta)^{\top}$ at
the r.h.s), and $\left[\partial (x,y,z)/\partial (\xi,\eta,\zeta)\right]^{-\top}$
is the inverse and transposed Jacobian matrix of the 
trans\-for\-mation (\refeq{eq:xyz}). 

\vspace{\baselineskip}
\noindent
\textbf{Remark (On softwares).} Although general-purpose programming
languages, such as FORTRAN and MAT\-LAB has been successfully applied to FEA
(see, for example, \cite[Chapter 12]{Bathe2006}, \cite{KwonB2000})
alternatively, one can use specialized software like ANSYS
(see~\cite{wiki:ANSYS}) or ADINA (which has the MITC9 shell element available
out-of-the-box, see~\cite{wiki:ADINA}), or, try software from other projects that
have more recently come out, such as FEniCS (see~\cite{wiki:FEniCS}), or,
Firedrake (see~\url{https:firedrakeproject.org}). 

\section{A Simple Model Problem}
\label{sec:SimpleModelProblem}
\begin{figure}[h!]
	\centering
	\includegraphics[width=80mm]{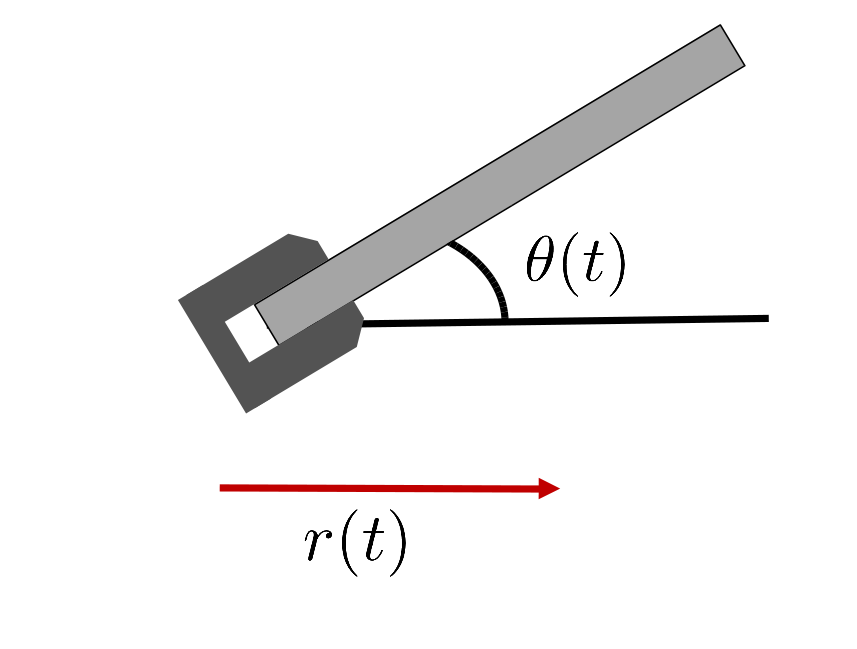}
	\caption{Schematic of model problem, showing rod, light grey, clamped at one end.}
	\label{fig:1DSchematic}
\end{figure}

Here we present a simple model problem to aid analysis. In order to do this we will consider a 1D analog to the problem presented in the earlier sections. We will consider a thin rod, clamped at one end, that is free to move and rotate in a single plane (see Figure \ref{fig:1DSchematic}).

\noindent The appropriate model to describe the deformation of the rod is the Euler-Bernoulli equation. As with the previous model for a 2D plate we will orient our axis with respect to the rod; with the origin being the clamped end of the rod; $x$ being the direction parallel to the undisplaced rod; $z$ being the direction perpendicular to the undisplaced rod, in the plane of motion of $r$ and $\theta$; and $y$ is the direction perpendicular to the rod, into the page in Figure \ref{fig:1DSchematic}. The Euler-Bernoulli equation reads
\begin{align*}
    EI\frac{\partial^4 w}{\partial x^4} = q - \rho \frac{\partial^2 w}{\partial t^2},
\end{align*}
\noindent where $E$ is the elastic modulus, $I = \iint z^2 dydz$, is the second moment of area, $w$ is the displacement of the rod in the direction perpendicular to the rod, $q$ is the load on the rod (the force applied to the rod in the perpendicular direction), $\rho$ is the density per unit length of the rod, and $t$ is time. $EI$ is often referred to as the flexural rigidity, and is the 1D analog to the bending stiffness, $D$, in the Kirchoff-Love equation. We must establish the form of $q$ due to the motion of the rod. We will omit the derivation here, as it is functionally the same as the derivation in the 2D plate case. The form that the load takes is
\begin{align*}
    q = \rho\left[\ddot{r} \sin(\theta) - g \cos(\theta) + \dot{\theta}^2 w - \ddot{\theta} x\right],
\end{align*}
\noindent where the terms are, moving left to right, the acceleration force, the force due to gravity, the Centrifugal force, and the Euler force. The appropriate boundary conditions for $w$ are
\begin{align*}
    w(x,0) &= \frac{\partial w}{\partial t}(x,0) = 0, \text{ initially undeformed and stationary,}\\
    w(0,t) &= \frac{\partial w}{\partial x}(0,t) = 0, \text{ clamped at one end,}\\
    \frac{\partial^2 w}{\partial x^2}(L,t) &= \frac{\partial^3 w}{\partial x^3}(L,t) = 0, \text{ free at one end,}
\end{align*}
\noindent where $L$ is the length of the rod. Whilst we are currently imposing $r$ and $\theta$, finding the stress for a given path, a piece of further work will be to find the path that minimises the stress. With this in mind we will give boundary conditions for $\theta$ and $r$
\begin{align*}    
    \theta(0) &= \theta_0, \quad \dot{\theta}(0) = 0, \quad \theta(T) = \theta_1, \quad \dot{\theta}(T) = 0,\\
    r(0) &= 0, \quad \dot{r}(0) = 0, \quad r(T) = R, \quad \dot{r}(T) = 0.
\end{align*}     
\noindent In other words, the rod is initially at rest at some specific angle, $\theta_1$, and ends at rest, at some later time, $T$, at position $R$ and angle $\theta_1$. Finally we have the form of the stress, $\sigma$,
\begin{align*}
    \sigma = -h E \frac{\partial^2 w}{\partial x^2},
\end{align*}
\noindent where $h$ is half the thickness of the rod in the $z$ direction. We now seek to non-dimensionalise with the following scalings
\begin{align*}
    x = L\hat{x}, \quad t = T\hat{t},\quad w = W\hat{w}, \quad r = R\hat{r}, \quad \sigma = \frac{hE}{L} \hat{\sigma}, 
\end{align*}
\noindent where $W$ is a characteristic lengthscale for the displacement. This gives us the following, dropping the hats for convenience,
\begin{align*}
    \lambda \frac{\partial^4 w}{\partial x^4} &= \mu \ddot{r} \sin(\theta) - \frac{1}{\mathrm{Fr}^2} \cos(\theta) + \dot{\theta}^2 w - \nu \ddot{\theta} x - \frac{\partial^2 w}{\partial t^2},\\
    \sigma &= -\frac{\partial^2 w}{\partial x^2},
\end{align*}
\noindent where $\lambda = \frac{EIT^2}{\mu L^4}$, $\mathrm{Fr} = \sqrt{W/gT^2}$, $\mu = R/W$, $\nu = L/W$ are our non-dimensional groups, together with the initial and boundary conditions
\begin{align*}
    \left.w\right|_{x=0} &= 0, \quad \left.\frac{\partial w}{\partial x}\right|_{x=0} = 0,\quad\left.\frac{\partial^2 w}{\partial x^2}\right|_{x=1} = 0,\quad\left.\frac{\partial^3 w}{\partial x^3}\right|_{x=1} = 0,\\
    \left.w\right|_{t=0} &= 0,\quad\left.\frac{\partial w}{\partial t}\right|_{t=0} = 0,\\
    \quad\theta(0) &= \theta_0,\quad\dot{\theta}(0) = 0,\quad\theta(1) = \theta_1,\quad\dot{\theta}(1) = 0,\\
   r(0) &= 0,\quad\dot{r}(0) = 0,\quad r(1) = 1, \quad \dot{r}(1) = 0.
\end{align*}
\noindent Now we will now simplify this further by assuming $w$ does not depend on $t$, $\mathrm{Fr}$ is large, and $\ddot{\theta} = \ddot{r} = 0$. This gives us
\begin{align*}
    \lambda \frac{\partial^4 w}{\partial x^4} &= \dot{\theta}^2 w,\\
    w &= \frac{\partial w}{\partial x} = 0 \text{ at $x=0$,}\\
    \frac{\partial^2 w}{\partial x^2} &= \frac{\partial^3 w}{\partial x^3} = 0 \text{ at $x=1$.}
\end{align*}
\noindent We can define $\beta = \left(\frac{\dot{\theta}^2}{\lambda}\right)^{1/4}$. The above problem only has non-trivial solutions if
\begin{align*}
    \cos(\beta) \cosh(\beta) + 1 = 0,\label{NonTrivSolCond}
\end{align*}
\noindent in which case the solution is
\begin{align*}
    w = A \left[\cosh(\beta x) - \cos(\beta x) + \frac{\cosh(\beta) + \cos(\beta_n)}{\sin(\beta) + \sinh(\beta )} \left(\sin(\beta x) - \sinh(\beta x)\right)\right].
\end{align*}
It is worth noting that the non-trivial solutions occur precisely when the beam is rotating at one of its natural frequencies, the resulting solution is an example of resonance.

\section{Conclusion and Further Work}
\label{sec:ConcFutureWork}
To continue with the work we have started, we understand that both the numerical implementation of the finite element approach that we proposed in Section \ref{sec:NumSim} and a deeper analysis of the simple model considered in Section \ref{sec:SimpleModelProblem} have to be done. The finite element analysis can also be done by using some commercial software, like ANSYS (\url{https:ansys.com}), with a comparison of results. It is clear that the commercial software will have some advantages in the beginning, but the implementation of {\sl ad hoc} models, for example, for the air resistance over the pieces, will perhaps be more difficult. A long list of alternatives to ANSYS can be found in Wikipedia in the article {\sl \href{https://en.wikipedia.org/wiki/List_of_computer_simulation_software}{List of computer simulation software}}.

This last-mentioned is one of the major issues that remain unsolved, i.e., how to model the air resistance. One possibility would be to couple the solid mechanics calculations with fluid mechanics calculations; doing so would make the problem more complete, but at the same time very complex. Furthermore, reasonable and simple {\it ad hoc} models, perhaps based on Asymptotic Analysis techniques, should exist or could be designed. In our opinion, performing numerical experiments as well as the laboratory experiments in this direction should help in addressing these challenges. 

The analysis of the simple problem stated in Section \ref{sec:SimpleModelProblem} would also be important. It would, for example, simplify the analysis of the role of the so-called {\it jerk}, or, the first-derivative of the acceleration, on the deformations. It would also be useful to understand the differences between the quasi-static case (time-dependent exterior forces, but no inertial forces) and the true-dynamic case (including the inertia). It is clear that in fast movements of heavy bodies, the inertial forces will be very relevant, but the important thing here is to know the limits of this reasonable simplification.

This analysis of the {\it jerk} would also help in the direction of the design of the geometries of trajectories, to avoid non-progressive changes in accelerations. The changes in the accelerations are precisely the {\it jerk}. This is a classical problem in railway design, and relation to their results should be discussed.

\section*{Acknowledgements}

T.B. \& O.B. acknowledge the support provided by the EPSRC Centre for Doctoral Training in Industrially Focused Mathematical Modelling (EP/L015803/1). J.R.P.~is supported by MINECO (Spain) grant MTM PGC2018-100928-B-I00. S.K. is supported by the grant Severo Ochoa SEV-2017-0718. J.S.-M. acknowledges partial support by MINECO (Spain) grant MTM2017-84214-C2-1-P. 

\bibliography{references}
\bibliographystyle{IEEEtran}

\end{document}